\documentclass{article}

\usepackage{arxiv}

\usepackage[utf8]{inputenc} 
\usepackage[T1]{fontenc}    
\usepackage{hyperref}       
\usepackage{url}            
\usepackage{booktabs}       
\usepackage{amsfonts}       
\usepackage{nicefrac}       
\usepackage{microtype}      
\usepackage{lipsum}		
\usepackage{graphicx}
\usepackage{natbib}
\usepackage{doi}
\usepackage{amsmath}
\usepackage{multirow}
\usepackage{array}
\usepackage{bm}

\newcommand{\method}{\textsc{RPCL}}
\newcommand{\cdmr}{\textsc{CDMR}}
\newcommand{\ccps}{\textsc{CCPS}}
\newcommand{\mecfour}{MEC$^4$}
\newcommand{\na}{\textemdash}

\title{Learning Robust Pair Confidence for Multimodal Emotion-Cause Pair Extraction}

\date{}

\author{
\href{https://orcid.org/0009-0009-0451-2162}
{\includegraphics[scale=0.06]{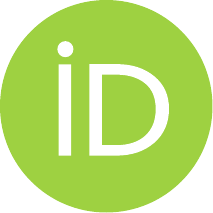}\hspace{1mm}Zhuangzhuang Pan} \\
Institute for Advanced Studies\\
Universiti Malaya\\
Kuala Lumpur 50603, Malaysia\\
\texttt{23078403@siswa.um.edu.my}
\And
\href{http://orcid.org/0000-0003-3045-9798}
{\includegraphics[scale=0.06]{orcid.pdf}\hspace{1mm}Ning Dong} \\
School of Information Engineering\\
Suqian University\\
Suqian 223800, China\\
\texttt{dongning@squ.edu.cn}
\AND
\href{http://orcid.org/0000-0003-2348-5082}
{\includegraphics[scale=0.06]{orcid.pdf}\hspace{1mm}Yingna Su} \\
School of Information Engineering\\
Suqian University\\
Suqian 223800, China\\
\texttt{suyingna@squ.edu.cn}
\And
\href{https://orcid.org/0009-0006-3559-4680}
{\includegraphics[scale=0.06]{orcid.pdf}\hspace{1mm}Yan Xia} \\
Digitization Department\\
Suzhou University of Technology\\
Suzhou 215500, China\\
\texttt{23072126@siswa.um.edu.my}
}

\renewcommand{\shorttitle}{Learning Robust Pair Confidence for MECPE}

\hypersetup{
  hidelinks,
  pdftitle={Learning Robust Pair Confidence for Multimodal Emotion-Cause Pair Extraction},
  pdfsubject={Multimodal emotion-cause pair extraction},
  pdfauthor={Zhuangzhuang Pan, Ning Dong, Yingna Su, Yan Xia},
  pdfkeywords={multimodal emotion-cause pair extraction, MECPE, robust pair confidence learning, pair-confidence learning, RPCL}
}

\begin{document}
\maketitle

\begin{abstract}
Multimodal emotion-cause pair extraction (MECPE) requires reliable pair confidence over candidate pairs. Existing pair scorers commonly use pair-level cross entropy over valid candidates, which treats links mostly independently. This leaves the relative confidence geometry among competing causes under-constrained, allowing gold pairs to stay close to hard negatives or rely on incidental non-gold context. We study this vulnerability as pair-confidence brittleness and propose \method{} (Robust Pair Confidence Learning), a training-only framework for pair-confidence learning. \method{} encourages pair confidence to be both discriminative and stable: gold pairs are separated from row-wise hard negatives through a confidence-difference margin constraint, and clean pair predictions are aligned with predictions from a corrupted view where non-gold contextual utterance representations are partially corrupted. The original clean pair scorer and decoding pipeline are used unchanged at inference time. On ECF, MECAD, and \mecfour{}, \method{} improves the three-seed mean Pair F1 over a matched base model by 2.58--2.83 percentage points in the full text-audio-video setting, and improves mean Pair AUPRC on all three datasets. Diagnostic analysis further shows larger gold-negative confidence gaps and lower margin-violation severity. These results suggest that explicitly shaping pair confidence is an effective training strategy for MECPE.
\end{abstract}


\section{Introduction}
\label{sec:intro}

Multimodal emotion-cause pair extraction (MECPE) in conversations aims to identify which utterances express emotions and which utterances cause them, forming emotion-cause pairs over a dialogue \citep{xia-ding-2019-emotion,wang-etal-2023-mmecp}. Compared with text-based emotion-cause pair extraction, it makes pair decisions inside a conversational structure where emotions, causes, speakers, and background turns are interleaved \citep{li-etal-2023-ecpec,jeong-bak-2023-conversational,hu-etal-2024-unimeec}. The relevant cause may be separated from the emotion by several turns, spoken by another participant, or supported unevenly by textual, acoustic, and visual cues \citep{wang-etal-2024-semeval,wu-etal-2025-emotion,li-etal-2023-ji,yu-etal-2025-beyond-verbal}. These properties make the task a structured pair decision problem: for a given emotion utterance, multiple candidate causes can be locally plausible, while only a small subset corresponds to annotated causal relations.

A common training practice is to supervise candidate pairs as positive or negative pair instances, often with cross-entropy-based objectives over valid candidates \citep{li-etal-2023-ecpec,cheng-etal-2023-acdmf,li-etal-2025-mmecpe,li-etal-2024-class}. This supervision is necessary, but it mainly evaluates each candidate through its own label. It does not directly enforce the relative confidence geometry needed when several causes compete for the same emotion. In difficult cases, a gold pair can remain close to non-gold candidates that share speaker, topic, temporal proximity, or multimodal affective evidence with the true cause \citep{wang-etal-2025-seg,ju-etal-2025-enhanced,ma-etal-2025-from}. This vulnerability is referred to as pair-confidence brittleness.

This paper studies multimodal emotion-cause pair extraction in conversations from the perspective of reliable pair confidence. A useful pair score should satisfy two complementary requirements. First, for a fixed emotion utterance, the score of a gold cause should be separated from the strongest non-gold alternatives for the same emotion. Second, the pair score should remain stable when contextual utterances outside annotated gold pairs are partially perturbed. These requirements complement recent progress in multimodal interaction, label constraints, memory-inspired modeling, and graph-based structure by directly shaping how a pair scorer allocates confidence among plausible links \citep{li-etal-2025-mmecpe,wu-etal-2025-emotion,liang-etal-2025-m3hg}.

To this end, this paper proposes RPCL (Robust Pair Confidence Learning), a training-only framework for pair-scoring emotion-cause models. RPCL adds no inference-time fusion module, decoder, or post-processing step. During training, it encourages two behaviors: gold pairs should stand apart from strong competing causes for the same emotion, and pair predictions should remain consistent when non-gold contextual evidence is partially corrupted. At inference time, the original clean pair scorer and the same decoding pipeline are used unchanged.

Evaluation is conducted on ECF, MECAD, and \mecfour{} using matched base scorers, identical input features, and unchanged decoding pipelines \citep{wang-etal-2023-mmecp,wu-etal-2025-emotion,liang-etal-2025-m3hg}. Overall, the contributions are:
\begin{itemize}
    \item We identify pair-confidence brittleness in MECPE and formulate reliable pair-confidence learning as a training problem beyond independent candidate-pair classification.
    \item We propose \method{}, a training-only framework that improves pair confidence by encouraging separation from strong non-gold alternatives and stability under label-preserving context perturbation.
    \item We verify the proposed mechanism through controlled comparisons and confidence diagnostics, showing improved pair extraction and better gold-negative confidence separation.
\end{itemize}

\section{Related Work}
\label{sec:related}

\paragraph{Structured Emotion-Cause Pair Extraction}
Emotion-cause pair extraction (ECPE) recasts affect analysis as link prediction between emotion and cause utterances rather than separate emotion/cause detection \citep{xia-ding-2019-emotion,li-etal-2023-ecpec}. Conversational extensions add speaker turns and dialogue context, while recent ECPE systems explore guided experts, commonsense generation, and semantic structure for more explicit causal reasoning \citep{jeong-bak-2023-conversational,wang-etal-2023-shark,yu-etal-2025-one,wang-etal-2025-seg}. These lines define the extraction space, but leave confidence geometry largely implicit.

\paragraph{Multimodal Emotion-Cause Pair Modeling}
Multimodal ECPE further binds causal links to textual, acoustic, and visual evidence, with ECF, SemEval-2024, MECAD, and \mecfour{} providing representative benchmarks \citep{wang-etal-2023-mmecp,wang-etal-2024-semeval,liang-etal-2025-m3hg,wu-etal-2025-emotion}. Existing systems strengthen pair modeling through holistic cross-modal interaction, causal prompting, memory-inspired aggregation, heterogeneous graphs, or LLM-enhanced generation \citep{hu-etal-2024-unimeec,cheng-etal-2024-mips,luo-etal-2024-nus,ju-etal-2025-enhanced,wang-etal-2024-observe}. They improve evidence encoding, whereas RPCL studies the confidence surface after scoring.

\paragraph{Training Objectives for Pair Reliability}
Several ECPE studies move beyond ordinary pair classification by making supervision more structurally consistent across emotion detection, cause detection, and pair extraction \citep{feng-etal-2023-joint,cheng-etal-2023-acdmf,hu-etal-2024-unifying}. Another line improves the training signal through stronger representations or sampling strategies for imbalanced candidate pairs \citep{hu-etal-2024-improving,su-etal-2024-hse-ttls,li-etal-2024-class}. Recent reliability-oriented studies further revisit confidence calibration, negative-sample regularization, and consistency under noisy views \citep{huang-etal-2026-revisiting,luo-etal-2026-supclap,he-etal-2026-rear}. However, these objectives regularize labels, tasks, examples, or representations rather than the row-conditioned confidence geometry in which a gold cause must outrank hard alternatives for the same emotion. RPCL adds this missing row-wise pressure and corrupted-context stability while preserving clean inference.

\section{Method}
\label{sec:method}

\subsection{Problem Formulation}
\label{subsec:problem}

Given a dialogue $D=\{u_i\}_{i=1}^{n}$, each utterance $u_i$ may contain textual, acoustic, and visual information. The task of multimodal emotion-cause pair extraction is to identify the set of emotion-cause pairs
\begin{equation}
Y=\{(i,j): u_i \text{ expresses an emotion and } u_j \text{ is its cause}\}.
\end{equation}
Let $\mathcal{V}\subseteq \{1,\ldots,n\}^2$ denote the valid candidate pair set under the adopted decoding scheme, and let
\begin{equation}
y_{ij}=\mathbf{1}[(i,j)\in Y], \qquad (i,j)\in\mathcal{V},
\end{equation}
be the pair label.

We build on a general multimodal ECPE backbone. For each dialogue, the backbone first produces a multimodal utterance representation $h_t$ for each utterance $u_t$. Based on these representations, it outputs emotion logits $z_i^e$ for utterance $u_i$, cause logits $z_j^c$ for utterance $u_j$, and pair logits $s_{ij}\in\mathbb{R}^{2}$ for each valid candidate pair $(i,j)\in\mathcal{V}$. The pair scorer can be viewed as a module that consumes the dialogue-level utterance representations and the candidate indices:
\begin{equation}
s_{ij}=f_{\mathrm{pair}}(\{h_t\}_{t=1}^{n},i,j).
\end{equation}
We denote the pair distribution and the positive pair confidence by
\begin{equation}
\bm{\pi}_{ij}=\operatorname{softmax}(s_{ij}), \qquad
p_{ij}=\bm{\pi}_{ij,1}.
\label{eq:pair-prob}
\end{equation}
Here, $p_{ij}$ is the confidence used by the pair scorer to decide whether $u_j$ is the cause of the emotion in $u_i$. The same pair-scoring interface is later used by the corrupted branch, where $\{h_t\}_{t=1}^{n}$ is replaced with the corrupted representations $\{\tilde h_t\}_{t=1}^{n}$.

\subsection{Overview of Robust Pair-Confidence Learning}
\label{subsec:overview}

We view multimodal ECPE as learning a structured \emph{pair-confidence surface} over valid emotion-cause candidates. Since all candidate pairs in a dialogue share the same conversational context, reliable pair confidence is shaped by two coupled factors: competition among alternative causes for the same emotion and stability when non-causal context is perturbed.

\begin{figure}[t]
\centering
\includegraphics[width=\linewidth]{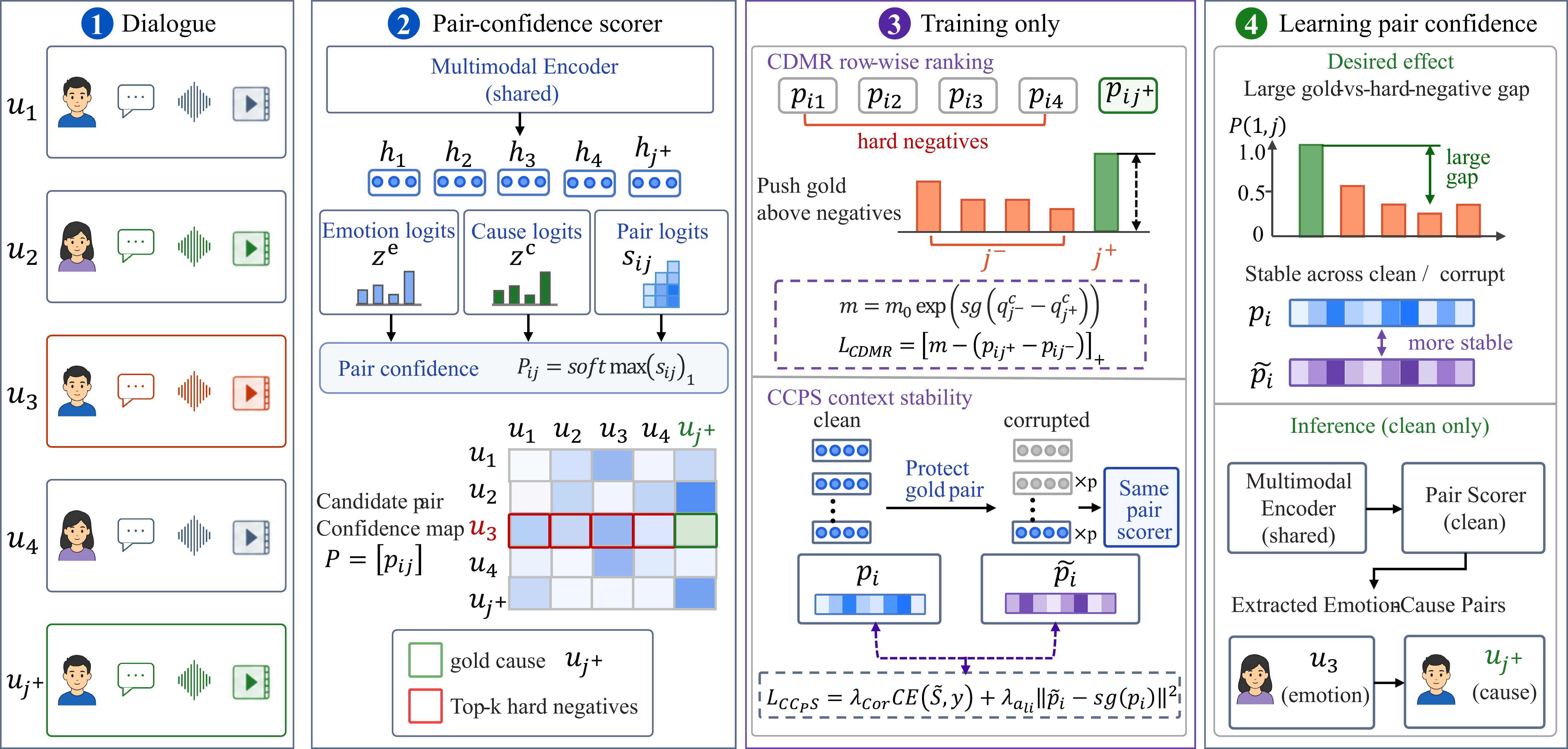}
\caption{Overview of RPCL. CDMR separates gold pairs from row-wise hard negatives, and CCPS aligns clean/corrupted predictions after protected context corruption.}
\label{fig:rpcl-overview}
\end{figure}

Standard pair-level cross entropy supervises each valid pair by its binary label, but it does not explicitly shape this row-wise confidence geometry and trains only on clean dialogues. Consequently, a gold pair may remain close to hard negatives in the same emotion row or depend on incidental non-gold context. \method{} addresses this with two training-only constraints: (i) \textbf{row-conditioned margin ranking}, which separates gold pairs from top-$k$ hard negatives within the same row, and (ii) \textbf{corrupted-context pair stability}, which preserves gold-pair evidence while perturbing non-gold utterances and aligning the resulting pair predictions. Both constraints act on the original pair scorer, and the inference pipeline remains unchanged. Figure~\ref{fig:rpcl-overview} summarizes the framework.

\subsection{Row-Conditioned Margin Ranking}
\label{subsec:cdmr}

We first make pair confidence discriminative within each emotion row. For an emotion utterance $u_i$, let
\begin{equation}
P_i=\{j:(i,j)\in\mathcal{V},\, y_{ij}=1\}, \qquad
N_i=\{j:(i,j)\in\mathcal{V},\, y_{ij}=0\},
\end{equation}
where $P_i$ and $N_i$ are the gold cause set and the non-gold candidate set for row $i$, respectively. The constraint is applied only to rows where both sets are non-empty.

Among all non-gold candidates, the most informative ones are those that the current model already considers plausible. We therefore mine the top-$k$ hard negatives according to the current pair confidence:
\begin{equation}
H_i=\operatorname{TopK}_{j\in N_i}(p_{ij}),
\label{eq:hard-neg}
\end{equation}
where $H_i$ contains the indices of the selected negatives. If fewer than $k$ negatives are available, all negatives are used. The $\operatorname{TopK}$ operation is used only to select negative candidates in the current forward pass. We do not back-propagate through the discrete selection itself.

For each gold cause $j^+\in P_i$ and hard negative $j^-\in H_i$, the model is encouraged to satisfy
\begin{equation}
p_{ij^+} - p_{ij^-} \ge m_{i,j^+,j^-}.
\end{equation}
The margin should be larger when the hard negative also appears cause-like. We use the cause classifier as a confidence signal. Let
\begin{equation}
q_j^c=\operatorname{softmax}(z_j^c)_1
\end{equation}
be the probability that utterance $u_j$ is a cause utterance. The adaptive margin is defined as
\begin{equation}
m_{i,j^+,j^-}
=
m_0 \exp\!\left(
\operatorname{sg}(q_{j^-}^c-q_{j^+}^c)
\right),
\label{eq:adaptive-margin}
\end{equation}
where $m_0$ is a base margin and $\operatorname{sg}(\cdot)$ denotes stop-gradient. Thus, the cause-confidence contrast determines the required pair-confidence gap, but the margin value itself does not back-propagate into the cause classifier.

This gives the Confidence-Difference Margin Ranking constraint:
\begin{equation}
\mathcal{L}_{\mathrm{CDMR}}
=
\frac{1}{|\Omega|}
\sum_{(i,j^+,j^-)\in\Omega}
\left[
m_{i,j^+,j^-}
-
\left(p_{ij^+}-p_{ij^-}\right)
\right]_+,
\label{eq:cdmr}
\end{equation}
where $[x]_+=\max(0,x)$ and
\begin{equation}
\Omega=\{(i,j^+,j^-): j^+\in P_i,\; j^-\in H_i\}.
\end{equation}
If $\Omega$ is empty in a mini-batch, we set $\mathcal{L}_{\mathrm{CDMR}}=0$.

The effect of Eq.~\eqref{eq:cdmr} is local and row-conditioned. It does not replace pair classification; rather, it focuses additional pressure on cases where the pair confidence is most likely to be brittle: gold pairs competing with high-confidence false causes for the same emotion utterance.

\subsection{Corrupted-Context Pair Stability}
\label{subsec:ccps}

The second constraint targets the stability of pair confidence under label-preserving context corruption. A model may classify a pair correctly in the clean dialogue but rely on incidental non-gold context to do so. To discourage this behavior, RPCL constructs a corrupted view that preserves annotated pair evidence while perturbing non-gold utterance representations.

Let $h_t$ denote the multimodal utterance representation of $u_t$ consumed by the pair scorer. We first identify utterances that participate in at least one gold pair:
\begin{equation}
G=\{t:\exists(i,j)\in Y,\; t=i \text{ or } t=j\}.
\label{eq:gold-utterance-set}
\end{equation}
Utterances in $G$ are protected. For each utterance outside $G$, we sample a Bernoulli corruption variable $r_t\sim\operatorname{Bernoulli}(\rho)$ and construct
\begin{equation}
\tilde h_t =
\begin{cases}
h_t, & t\in G,\\
(1-r_t)h_t, & t\notin G.
\end{cases}
\label{eq:corruption}
\end{equation}
Thus, a sampled non-gold utterance representation is zeroed out with probability $\rho$, while all utterances involved in annotated gold pairs remain unchanged. The corrupted dialogue view is used only during training.

Running the same pair scorer on the corrupted representations yields pair logits $\tilde{s}_{ij}$ and pair distributions
\begin{equation}
\tilde{\bm{\pi}}_{ij}=\operatorname{softmax}(\tilde{s}_{ij}).
\end{equation}
Because the gold pair evidence is protected, the original pair labels remain valid for the corrupted view. We therefore supervise the corrupted view with the same pair labels and, at the same time, align its pair distribution with the clean prediction:
\begin{equation}
\mathcal{L}_{\mathrm{CCPS}}
=
\lambda_{\mathrm{cor}}\,
\operatorname{CE}_{\mathcal{V}}(\tilde{s},y)
+
\lambda_{\mathrm{ali}}\,
\frac{1}{|\mathcal{V}|}
\sum_{(i,j)\in\mathcal{V}}
\left\|
\tilde{\bm{\pi}}_{ij}
-
\operatorname{sg}(\bm{\pi}_{ij})
\right\|_2^2 .
\label{eq:ccps}
\end{equation}
Here, $\operatorname{CE}_{\mathcal{V}}$ is the average pair cross-entropy over valid candidates, and $\lambda_{\mathrm{cor}}$ and $\lambda_{\mathrm{ali}}$ control the two parts of the constraint. The stop-gradient operation makes the clean prediction the reference distribution. Consequently, the corrupted branch is trained to preserve clean pair confidence, without forcing the clean branch to move toward a noisier corrupted prediction.

\subsection{Training Objective and Inference}
\label{subsec:training}

We train the backbone with the standard supervised extraction objective on the clean dialogue:
\begin{equation}
\mathcal{L}_{\mathrm{sup}}
=
\frac{1}{n}\sum_{i=1}^{n}
\operatorname{CE}(z_i^e,y_i^e)
+
\frac{1}{n}\sum_{j=1}^{n}
\operatorname{CE}(z_j^c,y_j^c)
+
\operatorname{CE}_{\mathcal{V}}(s,y),
\label{eq:sup}
\end{equation}
where $y_i^e$ and $y_j^c$ are the utterance-level emotion and cause labels used by the backbone, and $\operatorname{CE}_{\mathcal{V}}(s,y)$ denotes the average pair cross-entropy over valid candidate pairs.

The full RPCL objective is
\begin{equation}
\mathcal{L}_{\mathrm{RPCL}}
=
\mathcal{L}_{\mathrm{sup}}
+
\lambda_{\mathrm{row}}\mathcal{L}_{\mathrm{CDMR}}
+
\mathcal{L}_{\mathrm{CCPS}},
\label{eq:rpcl}
\end{equation}
where $\lambda_{\mathrm{row}}$ controls the strength of row-conditioned margin ranking, while the internal weights of $\mathcal{L}_{\mathrm{CCPS}}$ are defined in Eq.~\eqref{eq:ccps}.

All RPCL-specific operations are training-time constraints. At inference time, we use the original clean dialogue, the backbone pair scorer, and the same thresholding or decoding pipeline as the base ECPE model. No hard-negative mining, corrupted view, gold-utterance protection, or additional post-processing is introduced during inference.

\section{Experiments and Analysis}

\subsection{Datasets and Experimental Setup}

\subsubsection{Datasets}

We evaluate on three multimodal ECPE benchmarks: ECF, MECAD, and \mecfour{}. ECF is an English benchmark, whereas MECAD and \mecfour{} are Chinese benchmarks \citep{wang-etal-2023-mmecp,wu-etal-2025-emotion,liang-etal-2025-m3hg}. All three datasets provide text, audio, and video modalities under the multimodal emotion-cause pair extraction formulation, with split-wise statistics summarized in Table~\ref{tab:dataset-stats}. We report Pair F1 and Pair AUPRC on the test split.

\begin{table*}[h]
\centering
\footnotesize
\caption{Split-wise statistics for ECF, MECAD, and \mecfour{}. Avg. pair distance denotes the mean absolute utterance-index distance $\lvert i-j\rvert$ within annotated pairs.}
\label{tab:dataset-stats}
\begin{tabular}{lrrr rrr rrr}
\toprule
Metric & \multicolumn{3}{c}{ECF} & \multicolumn{3}{c}{MECAD} & \multicolumn{3}{c}{\mecfour{}} \\
\cmidrule(lr){2-4}\cmidrule(lr){5-7}\cmidrule(lr){8-10}
& Train & Valid & Test & Train & Valid & Test & Train & Valid & Test \\
\midrule
Conversations & 1,001 & 112 & 261 & 684 & 126 & 179 & 1,277 & 404 & 443 \\
Utterances & 9,966 & 1,087 & 2,566 & 7,516 & 1,168 & 1,832 & 17,513 & 5,541 & 6,051 \\
Emotion-cause pairs & 7,055 & 866 & 1,873 & 5,788 & 977 & 1,312 & 5,792 & 1,854 & 2,115 \\
Speakers & 265 & 48 & 105 & 413 & 86 & 115 & 213 & 122 & 142 \\
Avg. pair distance & 0.85 & 0.98 & 0.93 & 1.02 & 0.86 & 0.79 & 0.80 & 0.83 & 0.83 \\
\bottomrule
\end{tabular}
\end{table*}

\subsubsection{Experimental Settings}

The compared variants use the same datasets, splits, input features, validation-based threshold search, and decoding pipeline. Base denotes the same backbone and pair scorer trained only with $\mathcal{L}_{\mathrm{sup}}$, while \method{} adds the proposed \cdmr{} and \ccps{} constraints during training. Unless otherwise stated, results are averaged over three seeds: 42, 345, and 456. Thresholds are selected on the validation split to maximize Pair F1 and are then fixed for test evaluation.

Implementation details are grouped as follows. (i) For model inputs, we use RoBERTa-base for text, wav2vec 2.0 features for audio, and CLIP features for vision \citep{liu-etal-2019-roberta,baevski2020wav2vec,radford2021learning}, with a task-specific hidden size of 400. Dialogues are truncated to at most 35 utterances, and utterances are truncated to at most 512 tokens. (ii) For optimization, all models are trained for up to 30 epochs with early stopping patience 5, batch size 32, weight decay $10^{-4}$, gradient clipping at norm 1.0, learning rate $10^{-5}$ for RoBERTa, and $5\times10^{-5}$ for non-backbone parameters. (iii) For \method{}, we use one hyperparameter setting across datasets, modality settings, and random seeds: $m_0=0.05$, $k=8$, corruption probability $0.30$, $\lambda_{\mathrm{row}}=0.3$, $\lambda_{\mathrm{cor}}=0.75$, and $\lambda_{\mathrm{ali}}=0.2$. Experiments are implemented with PyTorch~2.10.0 and Transformers~5.7.0 on NVIDIA A100-SXM4-80GB GPUs.

\subsection{Main Results on Complete Multimodal Evidence}

Across the complete TAV setting, \method{} consistently improves controlled pair extraction over Base, with mean Pair F1 gains of +2.58 on ECF, +2.59 on MECAD, and +2.83 on \mecfour{}, together with threshold-independent AUPRC gains on all three datasets (Table~\ref{tab:main}). Because Base and \method{} share the same validation-based threshold search and inference pipeline, these gains are attributable to the training objective rather than decoding or operating-point changes. The largest F1 and AUPRC gains occur on \mecfour{}, where Base has the lowest F1, suggesting that the confidence constraints are most useful in the hardest evaluated setting.

\begin{table*}[t]
\centering
\footnotesize
\caption{Main text-audio-video (TAV) results. Values are mean Pair F1/AUPRC percentages over three seeds, with standard deviation shown as subscript; deltas are computed from the reported means.}
\label{tab:main}
\begin{tabular}{lrrrrrr}
\toprule
Dataset & Base F1 & \method{} F1 & $\Delta$ F1 & Base AUPRC & \method{} AUPRC & $\Delta$ AUPRC \\
\midrule
ECF   & 55.71\textsubscript{$\pm$0.41} & \textbf{58.29}\textsubscript{$\pm$0.62} & +2.58 & 54.83\textsubscript{$\pm$0.76} & \textbf{56.46}\textsubscript{$\pm$0.41} & +1.63 \\
MECAD & 49.90\textsubscript{$\pm$0.26} & \textbf{52.49}\textsubscript{$\pm$0.81} & +2.59 & 46.05\textsubscript{$\pm$0.39} & \textbf{48.28}\textsubscript{$\pm$0.38} & +2.23 \\
\mecfour{}  & 35.85\textsubscript{$\pm$0.53} & \textbf{38.68}\textsubscript{$\pm$0.83} & +2.83 & 28.02\textsubscript{$\pm$0.31} & \textbf{30.64}\textsubscript{$\pm$0.82} & +2.62 \\
\bottomrule
\end{tabular}
\end{table*}

\begin{table*}[t]
\centering
\footnotesize
\setlength{\tabcolsep}{2pt}
\caption{Standard training-objective controls under text-audio-video (TAV). Values are mean Pair F1/AUPRC percentages over three seeds; subscripts show standard deviation.}
\label{tab:objective-baselines}
\begin{tabular}{>{\raggedright\arraybackslash}p{4cm}rrrrrr}
\toprule
Variant & ECF F1 & ECF AUPRC & MECAD F1 & MECAD AUPRC & \mecfour{} F1 & \mecfour{} AUPRC \\
\midrule
Base & 55.71\textsubscript{$\pm$0.41} & 54.83\textsubscript{$\pm$0.76} & 49.90\textsubscript{$\pm$0.26} & 46.05\textsubscript{$\pm$0.39} & 35.85\textsubscript{$\pm$0.53} & 28.02\textsubscript{$\pm$0.31} \\
Fixed-margin ranking & 58.10\textsubscript{$\pm$0.75} & \textbf{57.73}\textsubscript{$\pm$0.98} & 51.93\textsubscript{$\pm$0.40} & 47.71\textsubscript{$\pm$0.33} & 38.13\textsubscript{$\pm$0.24} & 30.05\textsubscript{$\pm$0.46} \\
Utterance-dropout consistency & 57.24\textsubscript{$\pm$0.41} & 56.84\textsubscript{$\pm$0.42} & 51.49\textsubscript{$\pm$0.96} & 46.99\textsubscript{$\pm$0.03} & 38.50\textsubscript{$\pm$0.56} & 30.42\textsubscript{$\pm$0.37} \\
\method{} & \textbf{58.29}\textsubscript{$\pm$0.62} & 56.46\textsubscript{$\pm$0.41} & \textbf{52.49}\textsubscript{$\pm$0.81} & \textbf{48.28}\textsubscript{$\pm$0.38} & \textbf{38.68}\textsubscript{$\pm$0.83} & \textbf{30.64}\textsubscript{$\pm$0.82} \\
\bottomrule
\end{tabular}
\end{table*}

\subsection{Comparison with Standard Training Objectives}
\label{subsec:objective-baselines}
Against two conventional objective controls in the same TAV setting, \method{} gives the best Pair F1 on all three datasets and the best AUPRC on MECAD and \mecfour{}, while Fixed-margin ranking is the only exception on ECF AUPRC (Table~\ref{tab:objective-baselines}). Fixed-margin ranking keeps row-wise ranking but replaces the cause-confidence-aware adaptive margin with a fixed margin and removes corrupted-context stability, while Utterance-dropout consistency removes row-wise margin ranking, corrupts utterance-level representations without gold-utterance protection, and aligns clean/corrupted pair distributions without corrupted-view pair supervision. Both controls improve over Base, but their weaker cross-dataset profile suggests that adaptive row-wise separation and protected corrupted-context pair stability are complementary rather than reducible to either standard objective alone. The ECF AUPRC exception suggests fixed margins can sharpen ranking, although RPCL yields stronger balanced extraction.

\subsection{Comparison with Published Systems}
\label{subsec:published-comparison}

The published-system comparison provides contextual positioning rather than an isolated component test, because the compared ECPE and MECPE systems differ in architectures, modalities, input features, training protocols, and evaluation settings (Table~\ref{tab:published-comparison}). This caveat is especially relevant for \mecfour{}, where M$^3$F remains stronger under a different architecture. The selected systems cover heuristic and two-stage extraction methods, multimodal interaction and label-constraint models, graph-based methods, generative frameworks, and large-model-based approaches. We therefore use this comparison as broader literature context, while the controlled analyses below assess the effect of robust pair-confidence learning.

\begin{table*}[!t]
\centering
\begingroup
\footnotesize
\setlength{\tabcolsep}{4pt}
\caption{Published ECPE/MECPE comparison on official splits. Baselines use reported best modality settings; \method{} reports three-seed means.}
\label{tab:published-comparison}
\begin{tabular}{>{\raggedright\arraybackslash}p{4cm} >{\centering\arraybackslash}p{1.8cm} ccc ccc ccc}
\toprule
\multicolumn{11}{l}{\textbf{(a) Dataset: ECF}} \\
\toprule
\multirow{2}{*}{Method} & \multirow{2}{*}{Modalities} & \multicolumn{3}{c}{Emotion Extraction} & \multicolumn{3}{c}{Cause Extraction} & \multicolumn{3}{c}{Pair Extraction} \\
\cmidrule(lr){3-5}\cmidrule(lr){6-8}\cmidrule(lr){9-11}
& & P & R & F1 & P & R & F1 & P & R & F1 \\
\midrule
Heuristic~\citep{wang-etal-2023-mmecp} & T+A+V & 73.62 & 79.68 & 76.48 & 54.91 & 50.28 & 52.44 & 36.99 & 26.77 & 31.01 \\
MECPE-2S~\citep{wang-etal-2023-mmecp} & T+A & 76.91 & 81.68 & 79.17 & 67.25 & 73.91 & 70.27 & 57.13 & 50.34 & 53.48 \\
HiLo~\citep{li-etal-2025-mmecpe} & T+A+V & 75.46 & 83.66 & 79.35 & 65.72 & 80.31 & 72.28 & 51.78 & 59.69 & 55.45 \\
\midrule
\method{} & T+A+V & 79.59 & 77.14 & 78.32 & 69.28 & 76.39 & 72.65 & 52.20 & 66.00 & 58.29 \\
\bottomrule
\addlinespace[3pt]
\multicolumn{11}{l}{\textbf{(b) Dataset: MECAD}} \\
\toprule
\multirow{2}{*}{Method} & \multirow{2}{*}{Modalities} & \multicolumn{3}{c}{Emotion Extraction} & \multicolumn{3}{c}{Cause Extraction} & \multicolumn{3}{c}{Pair Extraction} \\
\cmidrule(lr){3-5}\cmidrule(lr){6-8}\cmidrule(lr){9-11}
& & P & R & F1 & P & R & F1 & P & R & F1 \\
\midrule
SHARK~\citep{wang-etal-2023-shark,liang-etal-2025-m3hg} & T & 69.30 & 67.02 & 68.14 & 64.18 & 66.36 & 65.24 & 49.02 & 42.87 & 45.74 \\
GPT-4o~\citep{liang-etal-2025-m3hg} & T & 71.41 & 63.69 & 67.33 & 65.03 & 69.26 & 67.08 & 39.68 & 41.77 & 40.70 \\
M$^3$HG~\citep{liang-etal-2025-m3hg} & T & 72.34 & 67.84 & 70.02 & 66.12 & 70.24 & 68.12 & 54.82 & 46.42 & 50.27 \\
\midrule
\method{} & T+A+V & 72.97 & 52.95 & 61.37 & 62.57 & 78.12 & 69.47 & 40.95 & 73.09 & 52.49 \\
\bottomrule
\addlinespace[3pt]
\multicolumn{11}{l}{\textbf{(c) Dataset: \mecfour{}}} \\
\toprule
\multirow{2}{*}{Method} & \multirow{2}{*}{Modalities} & \multicolumn{3}{c}{Emotion Extraction} & \multicolumn{3}{c}{Cause Extraction} & \multicolumn{3}{c}{Pair Extraction} \\
\cmidrule(lr){3-5}\cmidrule(lr){6-8}\cmidrule(lr){9-11}
& & P & R & F1 & P & R & F1 & P & R & F1 \\
\midrule
MECPE-2S~\citep{wang-etal-2023-mmecp,wu-etal-2025-emotion} & T+A+V & \na{} & \na{} & \na{} & \na{} & \na{} & \na{} & 33.42 & 25.34 & 28.82 \\
HiLo~\citep{li-etal-2025-mmecpe,wu-etal-2025-emotion} & T+A+V & \na{} & \na{} & \na{} & \na{} & \na{} & \na{} & 32.12 & 38.70 & 35.11 \\
M$^3$F~\citep{wu-etal-2025-emotion} & T+A+V & \na{} & \na{} & \na{} & \na{} & \na{} & \na{} & 47.01 & 42.76 & 44.79 \\
\midrule
\method{} & T+A+V & \na{} & \na{} & \na{} & \na{} & \na{} & \na{} & 29.67 & 55.68 & 38.68 \\
\bottomrule
\end{tabular}
\endgroup
\end{table*}

\subsection{Pair-Confidence Diagnostics}

\begin{figure}[t]
\centering
\includegraphics[width=0.95\linewidth]{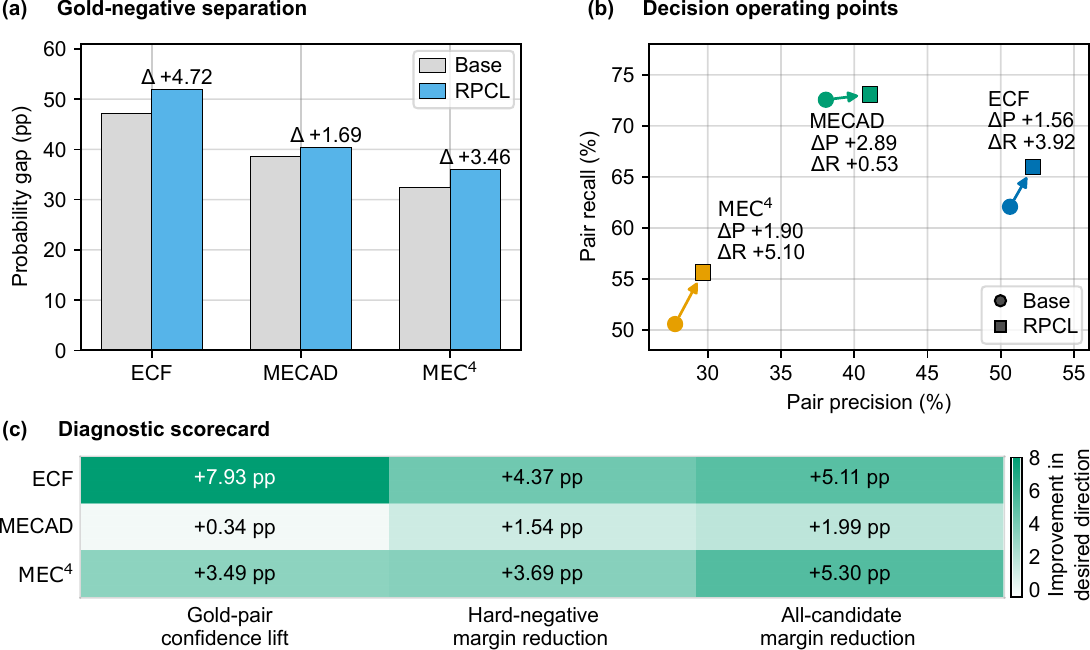}
\caption{Pair-confidence diagnostics under TAV. Results show gold-negative confidence gaps, precision-recall operating points, and desired-direction changes in gold confidence and margin-violation severity.}
\label{fig:pair-confidence-mechanism}
\end{figure}

The confidence diagnostics are consistent with the proposed mechanism rather than only the downstream F1 gains: \method{} increases the mean gold-minus-negative pair-probability gap by 4.72, 1.69, and 3.46 percentage points on ECF, MECAD, and \mecfour{}, respectively (Figure~\ref{fig:pair-confidence-mechanism}). The precision-recall movements differ by dataset, with ECF and \mecfour{} mainly gaining recall and MECAD mainly gaining precision, suggesting that \method{} does not simply bias the model toward more positive predictions. Gold-pair confidence rises on all three datasets, while hard-negative and all-candidate margin-violation severity decrease, which is consistent with better separation from competing candidates.

\begin{figure}[t]
\centering
\includegraphics[width=0.9\linewidth]{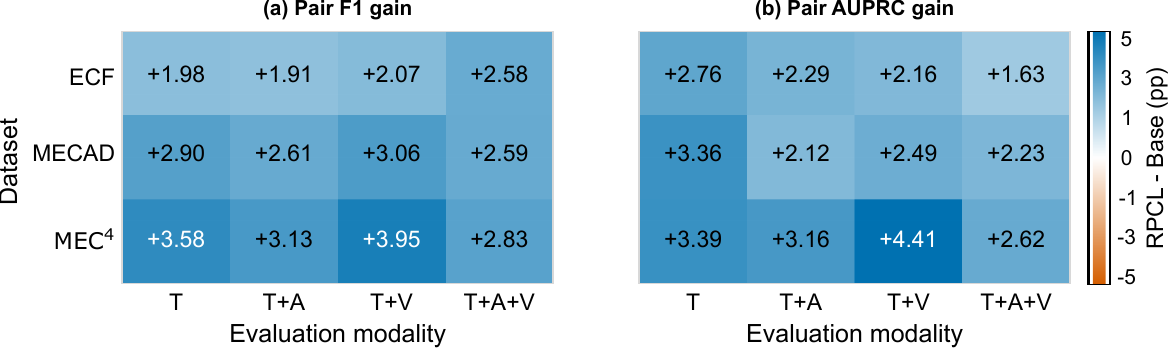}
\caption{\method{} gains over matched Base across modality settings. Cells show Pair F1 and Pair AUPRC changes in percentage points, averaged over three seeds.}
\label{fig:modality-gain}
\end{figure}

\subsection{Modality-Specific Behavior}

Across modality configurations, including text-only input without acoustic or visual evidence, the pair-confidence objective remains useful beyond the complete TAV setting: matched \method{}-Base gains are positive for Pair F1 and Pair AUPRC under T, T+A, T+V, and T+A+V (Figure~\ref{fig:modality-gain}). The T+A+V column corresponds to the complete-evidence setting in Table~\ref{tab:main}. The other columns show that the objective is not tied to a single input configuration, without isolating the relative contribution of each modality or the design of fusion modules.

\subsection{Ablation Study}

The ablations indicate that discriminative separation and stability both contribute to the final TAV performance: \cdmr{} alone and \ccps{} alone improve over Base on all datasets, and the full \method{} objective yields the best deltas on all reported metrics (Table~\ref{tab:ablation}). Partial removals show the same pattern internally: removing consistency or corrupted-view supervision weakens the \ccps{} path, while removing the adaptive margin or top-$k$ negative selection weakens \cdmr{}. Overall, removing either hard-negative separation or clean/corrupted stability reduces the \method{} gains.

\begin{table*}[t]
\centering
\footnotesize
\setlength{\tabcolsep}{3.2pt}
\caption{TAV ablation deltas over Base. Rows isolate or remove CDMR/CCPS components; positive values indicate percentage-point gains.}
\label{tab:ablation}
\begin{tabular}{lrrrrrr}
\toprule
Variant & ECF F1 & ECF AUPRC & MECAD F1 & MECAD AUPRC & \mecfour{} F1 & \mecfour{} AUPRC \\
\midrule
\cdmr{} only & +1.94 & +0.80 & +2.29 & +1.94 & +2.51 & +2.04 \\
\ccps{} only & +2.17 & +1.02 & +1.83 & +1.89 & +2.19 & +2.10 \\
\method{} w/o consistency & +2.39 & +1.27 & +2.21 & +2.11 & +2.17 & +2.10 \\
\method{} w/o corrupted CE & +2.24 & +1.18 & +2.08 & +1.97 & +2.05 & +2.03 \\
\method{} w/o adaptive margin & +2.33 & +1.36 & +2.42 & +2.08 & +2.46 & +2.29 \\
\method{} w/o top-$k$ negatives & +1.78 & +1.23 & +2.39 & +2.05 & +2.21 & +2.23 \\
\method{} & \textbf{+2.58} & \textbf{+1.63} & \textbf{+2.59} & \textbf{+2.23} & \textbf{+2.83} & \textbf{+2.62} \\
\bottomrule
\end{tabular}
\end{table*}

\section{Limitations}

\method{} is limited in three aspects. First, it is a training objective for pair-scoring backbones rather than a new multimodal encoder or decoder, so it may be complementary to stronger architectures. Second, the corrupted-context constraint uses representation-level perturbation and does not fully cover real-world noise such as ASR errors, missing visual frames, domain shift, or cultural variation. Third, because the task involves conversational multimodal emotion data, predictions should be interpreted only as annotated emotion-cause links, rather than as evidence of the underlying internal causes of emotion or as a basis for high-stakes decisions.

\section{Conclusion}

This paper addresses pair-confidence brittleness in multimodal emotion-cause pair extraction by framing pair extraction as reliable confidence learning over competing candidate pairs. We propose \method{}, a training-only framework that combines row-conditioned margin ranking with corrupted-context pair stability while leaving the original inference pipeline unchanged. Across ECF, MECAD, and \mecfour{}, \method{} consistently improves Pair F1 and Pair AUPRC over a matched base model, with diagnostics showing larger gold-negative confidence gaps and reduced margin-violation severity. These results show that explicitly shaping the pair-confidence surface, rather than only enriching representations or decoders, provides an effective and lightweight strategy for multimodal ECPE.

\bibliographystyle{unsrtnat}
\bibliography{references}

\end{document}